# Exiting the Simulation: The Road to Robust and Resilient Autonomous Vehicles at Scale

Rick Chakra

*Abstract*—In the past two decades, autonomous driving has been catalyzed into reality by the growing capabilities of machine learning. This paradigm shift possesses significant potential to transform the future of mobility and reshape our society as a whole. With the recent advances in perception, planning, and control capabilities, autonomous driving technologies are being rolled out for public trials, yet we remain far from being able to rigorously ensure the resilient operations of these systems across the long-tailed nature of the driving environment. Given the limitations of real-world testing, autonomous vehicle simulation stands as the critical component in exploring the edge of autonomous driving capabilities, developing the robust behaviors required for successful real-world operation, and enabling the extraction of hidden risks from these complex systems prior to deployment. This paper presents the current state-of-the-art simulation frameworks and methodologies used in the development of autonomous driving systems, with a focus on outlining how simulation is used to build the resiliency required for real-world operation and the methods developed to bridge the gap between simulation and reality. A synthesis of the key challenges surrounding autonomous driving simulation is presented, specifically highlighting the opportunities to further advance the ability to continuously learn in simulation and effectively transfer the learning into the real-world – enabling autonomous vehicles to exit the guardrails of simulation and deliver robust and resilient operations at scale.

## I. Introduction

In 1965, Ralph Nader rose to prominence by taking on General Motors (GM) in a David and Goliath manner. In his book, *Unsafe at Any Speed*, he charged GM and the automotive industry as a whole for failing to ensure the safety of the automobiles they produced [1]. Nader's charge was that this was not simply a failure of engineering, but also a failure of imagination. This charge prompted a transformation of the automotive industry, with the introduction of innovative technologies like seat belts, air bags, anti-lock brakes, and updated regulations surrounding the verification, validation, and overall safety of the vehicles being produced for the public.

While the safety of automotive transportation has significantly increased over the past half-century, the fundamental nature of driving remains the same, as the driver behind the wheel has not changed. There is a limit to how much protection can be put around an imperfect system governed by human limitations and biases. As evidence of this, there were nearly 37 thousand people killed in the United States in motor-vehicle accidents in 2018, with over 2.7 million injured [2]. More than 90% of these crashes were caused by human error. In comparison, only 2% were due to vehicle failures [3].

From an economic standpoint, these motor-vehicle crashes are estimated to exceed $800B in economic and social cost in the US [4]. In addition to the risk of injury and death, the exponential growth of personal mobility has led to an increase in congestion, with commuters spending nearly 7 billion additional hours waiting in traffic, with the cost of congestion reaching over $150B in 2015 [5]. The associated health cost of traffic congestion, calculated based on the premature deaths due to pollution inhalation, was estimated to exceed $20B in a 2010 study [6].

The magnitude of these figures is largely attributable to the nature of human-driven transportation. The questions surrounding mechanical and technological safety posed in the '60s have now transitioned to the limitations and imperfections of human drivers. Humans are limited in their ability to process high volume and high velocity information about their environment and imperfect in identifying the control thresholds of their vehicle. In addition, humans are prone to bad decisions, influenced by emotions of fear and anger, and present a threat to themselves and others on the road when driving impaired, tired, or distracted [7]. These accident metrics and limitations present an opportunity for not just automotive engineering innovation, but also in re-imagining what personal mobility looks like.

Autonomous vehicle (AV) technology has shown the potential to mitigate many of these crises by eliminating the mistakes that human drivers make [8], [9], while presenting new opportunities to improve transportation efficiency, reduce pollution, and increase social mobility [10]. From its initial developments, AV technology has been built on the premise that transitioning the driving task from humans to machines could potentially solve many of the social crises surrounding mobility and transportation. Starting with Carnegie Mellon's NAVLAB vehicle in the late 80s, AV technology demonstrated the ability to perform lane-following maneuvers using camera images [11] – a key task in human driving. Expanding on these basic maneuvers, the PROMETHEUS project, a multi-organization research effort, significantly advanced the perception capabilities and showcased the ability of extended autonomous driving, which included lane changes and tracking of other vehicles [12], [13]. The early work presented in [11] and [13] set the path for more advanced AV technology and hinted at a potential future where this technology could surpass human driving capabilities.

The 2004 and 2005 DARPA Grand Challenges [14] and the 2007 DARPA Urban Challenge [15] propelled the research forward, as several AVs were able to successfully traverse a range of desert and urban terrains, effectively highlighting the transition point of AV technology from concept to reality. In parallel, the automotive industry began introducing Advanced Driver Assistance Systems (ADAS) like adaptive cruise control (ACC), lane keeping, and automatic parking, which incrementally paved the way for the development of fully autonomous vehicles, able to operate without any human intervention [10]. These systems have shown the ability to operate in limited and structured domains, and significant effort is now being applied to enable them to transition to the unstructured and stochastic nature of urban driving [16]. Leading AV companies like Tesla [17], Waymo [18], Uber [19], Aptive [20], Voyage [21], and Comma AI [22] have all showcased their latest autonomous driving technology but continue to highlight the key challenges that need to be solved [23]. AVs are not just an augmentation of existing technology, but rather a rewrite of the personal mobility architecture and the verification and safety standards that go along with it.

The driving task exists in a highly complex, unstructured, and integrated ecosystem, which requires a generalization of knowledge to successfully operate across the long tail of real-world driving scenarios. While AV technology has the potential to mitigate many of the existing driving issues and capitalize on new opportunities, the capability of humans to effectively handle new scenarios without scenario-specific guidelines or references, and with much less sensing and processing capabilities, cannot be downplayed. AVs may be able to identify, classify, and predict the motion of dozens of objects in parallel, while actuating steering and acceleration inputs at precise control thresholds, but they still lack the ability to do what humans can do – truly understand a scene and incrementally transfer that understanding from one scenario to another. These types of "intelligence" gaps may lead to AVs performing worse than humans in many cases [24]. In addition, AVs may present serious new risks like unknown and unexpected behaviors when presented with edge cases that have not been encountered before, as well as abnormal actions when dealing with malicious inputs intended to trick the AV perception system [25], [26]. In contrast, these scenarios would be easily recognizable by humans. Public examples of these failures [27]-[33] underscore the importance of understanding how AVs will behave when presented with these types of scenarios, without actually knowing the exact scenario. These types of accidents damage the ability to deploy scalable AV technology, given that social acceptance and investment is dependent on the measure of safe mileage driven and the associated increased level of trust [32]. Therefore, it is critical to be able to verify whether the whole system can safely operate across a wide range of interactive scenarios, especially involving risks to human-life. This validation process must take into consideration the critical interfaces between components of the AV systems, given that flaws in complex AV architectures routinely occur at subsystem interfaces and can introduce unexpected behaviors [34]. Understanding the operational resiliency and capabilities of an autonomous vehicle will look more like an IQ test [35], in comparison to the well-defined and standardized safety and performance testing of today's vehicles – which can be understood, measured, and repeated to assure engineers, regulators, insurers, and consumers of a vehicle's capability.

Although AV sensor and computing technology are decreasing in both size and price [36], and perception capabilities are maturing to human-level performance on narrow computer vision tasks like segmentation and object recognition [37], the solution of robust and resilient autonomous vehicles at scale is not just a matter of waiting for the technology. The technological advancements are a prerequisite to full autonomy and not the key to solving the core problems required to enable robust autonomous driving capabilities [16]. Even with the magnitude of time and money being invested in tuning the highly engineered AV systems and subsystems, the problem of autonomous driving is still far from being solved [38]. As highlighted by Waymo – they may be 90% done, but still have 90% to go [23].

To address the core challenges, the academic field as well as industry have turned their focus to simulation. A recent study argued that AVs "cannot simply drive their way to safety," given that they would need to drive "hundreds of millions of miles and, under some scenarios, hundreds of billions of miles to create enough data to clearly demonstrate their safety" [7]. Furthermore, leveraging real-world driving to assess AV performance across constantly changing environments and edge cases would be prohibitively expensive and dangerous. Alternatively, leveraging formal verification methodologies to verifying the "correctness" of ADS algorithms and driving policies is difficult due to a lack of a formal definition of "correctness" [34], [39]. Unlike real-world testing, which is both limited and impractical in the ability to provide the scope and scale of testing scenarios [7], simulation allows engineers to vary the parameters of interest and accelerate the evaluation process in a controlled environment. The need for simulation to extract hidden risks and explore optimal design and performance is not a new discovery – some of the earliest work in this field highlighted this need [40]. Given that there is currently no standard system for AV testing, simulation will play a key role in the verification and validation of AVs. Effective simulation is not purely a technical silver bullet for automotive or technological problems with autonomous driving, it must be an interdisciplinary solution [41].

The motivation of this paper is to survey the current state-of-the-art simulation techniques used in autonomous driving and distill the key frameworks, technologies, and methodologies that have been developed to investigate AV capabilities, uncover hidden risks, and ultimately increase the robustness of AVs as these systems exit the guardrails of simulation and transition to the real world. The remainder of this paper is organized into four sections. Section II provides a high-level outline of AV architecture and system components and introduces how deep learning is being used to enable AV capabilities. Section III highlights the current verification and validation guidelines in development, testing, and certification of AVs. Section IV provided a survey of the key simulation frameworks and methodologies leveraged to explore AV design, extract hidden risks, and enable an effective transfer of learning from simulation to reality. Section V frames the key problems in this space and proposes a high-level roadmap for future research. The final section summarizes the paper and provides closing remarks.

## II. AV ARCHITECTURE AND SYSTEM COMPONENTS

### A. Background

The Society of Automotive Engineers has defined six levels to categorize AV technology, ranging from "Level 0 - No Automation" to "Level 5 - Full Automation" [42]. Most of the advanced driver-assistance systems (ADAS) available in production cars today, like lane keeping and automatic cruise control, are classified as "Level 1 - Driver Assistance" systems which only control lateral or longitudinal trajectories and the driver maintains full responsibility of the driving task. In contrast, the leading AV companies today are pursuing "Level 4 - High Automation" – transitioning the driving responsibility to the machine which operates in full autonomy under specific conditions. The path to Level 4 automation varies, as some companies are pursuing a strategy of deploying directly with Level 4 capabilities [18], while others are following a path of incremental improvements [17]. To enable this higher level of driving automation, AVs have become highly engineered and complex machines, leveraging deep learning (DL) to transition the driving task from human to machine [43]. The full automation of the driving task requires AVs to leverage a pipeline of integrated hardware and software subsystems to create a driving policy, defined as the mapping between sensor inputs and actuator outputs [44]. While the specific architectures may vary, all share the same three core elements: sensors to perceive the environment and measure the ego-vehicle movement, on-board computers for planning and prediction, and actuators for vehicle control [36]. These core capabilities can be summarized as Sense, Plan, and Act.

### B. Sense: Perception, Localization, and Sensor Fusion

The ability for an AV to perceive its local environment, including the identification and classification of static and dynamic objects, as well as their positioning relative to the vehicle, is one of the most critical and challenging components of the AV pipeline. The core perception task is achieved through the use of monocular and stereo cameras, as well as range sensing devises like RADAR and LiDAR [16]. In combination with satellite positioning systems like GPS, GNSS, and GLONASS, and inertial navigation systems (INS), AVs are able to combine their perception data with positional data to estimate the vehicle's pose and motion [45]. Data from these various sensors are merged through a sensor fusion process, which produces data-rich scene representations used to detect key features, such as road boundaries, traffic signs, and other vehicles. In addition, these scene representations integrate with downstream planning and control components to enable the navigation and interaction required to traverse complex driving scenarios.

To generate these detailed scene representations, DL algorithms are leveraged to detect the drivable areas and identify the locations, velocities, and potential future states of the surrounding objects [36]. Given the stochastic nature of the driving environment, the key role of an effective perception system is to filter out the environmental noise while preserving the information needed for robust planning and control procedures [38]. In addition, high definition (HD) maps play a key role in the perception pipeline for some AV architectures [18]. They consist of annotated geometric, semantic, and prior map layers which offer a predefinition of the static driving environment and present derived information about the dynamic objects that a vehicle may encounter. These capabilities serve as hints to the AV pipeline. Without a semantic map, the static information in the environment would need to be continuously perceived by the AV sensor suite and processed by the CPUs; therefore, this is a powerful tool in the precomputing and offloading of static information in the AV perception layer [36]. In contrast, there are other successful AV applications that do not leverage HD maps [17], [22]. Instead, vision-based perception is heavily utilized to execute the end-to-end driving process, including identifying traffic behaviors, approximating intersection structures, detecting static objects, and inferring scene-specific information.

Given that this sensory component controls the collection of data and the recognition required to develop an understanding of the scene, it is critical in the overall success of the AV driving policy. Any errors or faults in this layer can drastically impact the ability of the subsequent layers to properly function. Due to the non-linear nature of AV systems, small errors in perception can result in catastrophic failures in planning and control. As evidence of this, many of the fatal accidents involving AVs have been due to perception system failures [28]-[31]. Therefore, the accuracy and resilience of this layer are critical measures in the ability to deploy robust AVs at scale.

The surveys presented in [36] and [45] provide a detailed overview of the sensors and systems involved in perception and localization. Furthermore, the work in [46] expands on the various DL techniques involved in this layer.

### C. Plan: Prediction and Planning

The data-rich scene representations produced by the perception system are fed to the path prediction and planning components to execute the mission-critical decision-making process, which selects the optimal vehicle trajectory that ensures safety, comfort, and route efficiency [47]. The planning component typically leverages a hierarchical framework which consists of layers responsible for mission planning, behavioral planning, and motion planning [36]. The mission planner is responsible for the high-level objectives, like defining the planned route to reach the target destination. The behavioral planner utilizes the sequence of road segments specifying a selected route and makes the decisions required to interact with other agents while following the rules and social norms of the road, resulting in local objectives like lane changing, overtaking, or intersection navigation. This layer must also be able to handle the uncertainties related to the intentions and behaviors of other traffic participants [48]. The motion planner integrates with the scene representation and local objectives to generate trajectories that enable the vehicle to navigate in a safe, comfortable, and efficient manner [36].

To be able to select safe and adaptable trajectories across the wide range of scenarios an AV will encounter, especially in complex urban environments, the planning layer must be able to reliably predict the future behaviors of the other dynamic agents present in the environment. To do this, it considers the predicted trajectories of both motorized and non-motorized traffic participants to filter out potentially problematic paths. The behavior of the other traffic participants is predicted through motion-modelling techniques. These techniques range from physics-based and maneuver-based models, which do not consider the interaction between vehicles, to interaction-aware models that enable the utilization of inter-vehicle relationships in the prediction of dangerous trajectories [47].

In addition, the integration of vehicle-to-vehicle (V2V) and vehicle-to-infrastructure (V2I) communication technologies have the potential to expand the look-ahead horizon of AVs and enhance their ability to predict the behaviors of other dynamic agents [47].

The surveys presented in [48], [47], and [36] provide a comprehensive outline of the systems and techniques utilized in the planning and prediction layer of the AV architecture.

*D. Act: Integration and Control*

The controller is responsible for selecting the appropriate actuator inputs required to execute the planned trajectories produced by the planning system. While the upstream layers constitute the cognitive layers of the AV, the control components must convert the intention of the AV into real-world forces and energy. The motion control of a vehicle can be classified as two integrated tasks – lateral motion, which is controlled by the steering, and longitudinal motion, which is controlled through the acceleration and deceleration of the vehicle [10]. One of the standard models for motion planning is Model Predictive Control (MPC) [48], which seeks to optimize the trajectory of the vehicle over a forward time horizon, called the prediction horizon. In addition to executing the planned trajectory, the controller is also responsible for tracking the errors generated during the driving process, which are delivered back through a feedback loop to the planning components for error adjustments and corrections.

Comprehensive surveys of the control techniques leveraged in the AV architecture are detailed in [36] and [48].

*E. Autonomous Vehicle Architectures*

The layers and subsystems responsible for the sense-plan-act pipeline can be organized into various AV architectures. The traditional approach is the modular pipeline. It consists of a structured sequence of separate components which take in the sensory inputs, process them through the various prediction and planning subcomponents, and produce output trajectories that are fed to the system actuators for execution and feedback [37]. The other approach is the end-to-end architecture, which generates the output actions directly from the sensory inputs [40], [49], [50].

While modular pipelines have been the standard approach to achieving self-driving capabilities, the end-to-end approach, which is able to go from input pixels to output actions, has the potential to outperform them. End-to-end architectures enable the driving policy to bypass the human-selected optimization criteria, like lane line detection, and instead focus on optimizing the overall system performance [49]. In turn, this type of system learns the internal representations of the complete driving task, instead of specific detection, planning, and control objectives. There is significant effort being applied to improve the end-to-end approach in the AV space [51], as it has the potential to optimize all driving steps in parallel, leading to a smaller network with higher performance in obscure and challenging conditions [49].

Codevilla et al. [50] present how this end-to-end architecture can be expanded to include conditional high-level command inputs. These inputs create a communication channel with the network controlling the car, which can be used to specify the navigational intent of human occupants and guide the vehicle through complex environments. In addition, the end-to-end approach can be integrated with the traditional modular approach to form a hybrid architecture. Muller et al. [38] present how an end-to-end planning and control policy can be encapsulated between modular perception and control layers to achieve a hybrid end-to-end architecture. This type of architecture leverages modularization and abstraction between layers to enable the integration of end-to-end capabilities with traditional modular components.

From a connectivity perspective, these systems can take the form of ego-only systems or connected systems [37]. While there are no fully-connected AVs in operation today, V2V and V2I connectivity is expected to increase with the growth of AV deployment [52]. This integration enables the local AV architecture to expand to an AV ecosystem. As a result of this connectivity, many of the critical shortcomings and challenges of autonomous driving such as sensing range, limited visibility, and computational limitations can potentially be mitigated [37].

### III. AV DESIGN AND TESTING GUIDELINES

*A. Background*

Given that the AV architecture leverages integrated DL models to perceive their environment, generate the detailed scene representation, and select optimal vehicle trajectories, AVs are highly probabilistic in nature. Therefore, the outcome of an AV's interaction with the stochastic driving environment cannot be defined with absolute certainty, as an AV can present different behaviors in real-world operation than what was displayed during testing and certification [41]. This variability in the handling of uncertain interactions including the presence of static objects, pedestrian, and other vehicles, across a wide range of environments, creates a challenge in effectively verifying, validating, and certifying these systems. In addition, AVs are classified as safety and security-critical systems, given the risk to human life, and must be developed and tested in a manner that follows rigorous and systematic guidelines to expose any undesired behaviors, flaws, and vulnerabilities [53].

One of the key gaps in the ability to design, verify, and certify AVs for deployment is the lack of a detailed and testable definition of AV intelligence [54]. Furthermore, the DL components which define the intelligence and capabilities of these systems can consist of thousands of neurons and millions of parameters. This poses a new challenge to systematic and automated testing of large-scale AI systems deployed in interactive real-world environments [55]. Without a detailed and testable definition of AV intelligence, the ability to perform tests that produce quantitative, repeatable, and comparable results at scale remains one of the key challenges in developing and deploying AV technology onto public roads [54]. In contrast, human drivers undergo minimal testing that consists of road signs, traffic questions, and a simple driving assessment, to ensure they meet the threshold for operating a vehicle. The sufficiency of such a simple test is rooted in the ability to trust the generalizability and transferability of human intelligence, and the human ability to easily recognize different traffic situations, make the appropriate decisions, and execute the corresponding actions required to safely interact with the dynamic environment under accepted safety levels. That same ability does not exist for AVs. Therefore, the key question for deploying AVs at scale is, "how can we prove an autonomous vehicle is capable to drive in live traffic?" [54].

The existing AV intelligence testing approaches consist of scenario-based testing and functionality-based testing. Scenario-based testing is similar to the DARPA challenges, where an AV is required to interact with and navigate a designated region with a collection of scenarios in a safe, legal, and effective fashion. On the other hand, functionality-based testing focuses on validating the subsystems of the AV including sensing and perception functionality, prediction and planning functionality based on the perceived environment, and the control and action functionality to execute the AV decisions. Both of these existing methods have critical shortcomings. Scenario-based testing limits the evaluation of the AV intelligence to external measures of performance without directly and quantitatively assessing the intelligence. In addition, the available traffic scenarios considered for testing are always limited, given the impossibility to design and present all possible scenarios. Therefore, validation techniques must leverage a limited prioritization framework to model the intelligence of the AV from a limited number of tests. While functionality-based testing enables a quantitative evaluation of the subsystems within a specifically designed test framework, it is limited to the validation of separate and independent functionalities, without a comprehensive and integrated vehicle intelligence test to reliably evaluate the AV system components and the key integrations between them [54].

*B. Key Guidelines and Requirements*

Although more than half of the state in the US have introduced legislation to enable the development, testing, and deployment of autonomous vehicles [56], there still does not exist a standard system for AV testing [41]. In the presence of this gap, there has been a collection of safety requirements, testing frameworks, and deployment guidelines that have been published which guide the development and testing of AVs. In terms of the existing international standards governing vehicle systems, ISO26262 is the standard automotive framework for ensuring functional safety and mitigating system failures [57]. As highlighted in [41], ISO26262 relies on a human driver ultimately responsible for safety. Therefore, while this standard may be sufficient for ADAS applications where the human maintains control of the driving task, significant changes would be required to handle the transition of the driving task to the AV. Another automotive industry standard, ISO/PAS 21448, expands the requirements on "Safety of the Intended Functionality" (SOTIF) and defines safety as the absence of unreasonable risk [58]. As a similar limitation to ISO26262, ISO/PAS 21448 is focused on providing guidelines that assess hazards arising from SAE Level 1 and Level 2 systems but could potentially be expanded for higher levels of driving automation [59].

From a federal perspective, NHTSA published an expansive research report outlining a framework for AV validation. These guidelines describe the attributes that define the Operational Design Domain (ODD) for an AV system and outline the preliminary tests and evaluation measures that can be applied to validate the performance within a specific ODD and support the assessment required for safe deployment [60]. In addition, this report stresses the importance of simulation in validating various AV components and determining the overall safety of an AV prior to deployment. As additional guidance, NHTSA also recently published "A Vision for Safety 2.0" [53], which serves as additional AV testing guidance and outlines the importance of ensuring that AVs have the ability to react to the behaviors of other road users and effectively avoid or mitigate crashes.

In addition to the federal framework, the RAND Corporation published an extensive report proposing a framework for defining and measuring AV safety. They highlight that the standard measure of actual crashes and dangerous driving incidents is dependent on exposure to a wide range of scenarios over many miles for it to be useful, which is rare for AVs at this stage of development. They also stress the statistical challenges in comparing performance between AVs and human-driven vehicles, given that comparable data aligned across ODD is difficult to identify. Furthermore, they present the comparison between the evolution of AV safety and aviation safety and highlight the fact that carriers in the aviation field cooperate rather than compete on safety, given that a single aviation failure impacts the whole industry. This type of coordination and sense of shared fate is still lacking within the AV industry. Furthermore, they propose three key recommendations: (1) a focus on safety in addition to performance in the early stages of development, (2) a formal protocol on simulation demonstrations that facilitate safety assessments and comparisons across different AVs, and (3) outcomes of crashes or the absence of crashes should be evaluated as case studies to prevent statistics-based safety determinations that cannot be supported by sufficient exposure [61].

In addition to these guidelines, Underwriters Laboratories, a global safety certification company, also published a report which defines the standards that address the safety principles of fully autonomous vehicles, with a focus on SAE Level 3 to Level 5 technology. Their standard outlines a method to demonstrate AV safety through a "safety case approach" that is both goal-based and technology-agnostic. The key purpose of this standard is to recognize that transportation systems will always entail a level of risk. Therefore, the process for evaluating AV systems must focus on acceptable safety, where an AV is able to avoid unreasonable risks, and not absolute safety across all risks [62]. Furthermore, a group of international vehicle manufacturers and electronics system suppliers issued a summary of autonomous vehicle design and testing methodologies [63]. Similar to the "safety case approach" introduced in [62], the approach in [63] distills the key challenges in demonstrating the safety of Level 3 and Level 4

AVs and the impossibility of testing every imaginable driving scenario. To address these limitations, extensive use of simulation to validate the integrated AV architecture and its various subsystems is proposed [63].

*C. AV Design & Testing Projects*

Given the various requirements and frameworks that serve as guidelines for AV design and testing, there have been a few key testing consortiums that have formed to leverage testing-at-scale for AV verification and validation. The AdaptIVe project consisted of 28 partners across 8 European counties who worked together to collaboratively develop and test AV technologies and define the various evaluation scenarios required for scalable and repeatable testing [60]. The PEGASUS testing project was formed with 17 partners, including OEMs, Tier 1 suppliers, and scientific institutes. One of the key goals of this program was the identification and generation of testing scenarios at different levels of abstraction, with an objective of implementing these tests through a mix of simulation and real-world testing. In addition to the test scenario generation, the PEGASUS group sought to identify formal performance metrics for the techniques developed [60].

In addition to these testing projects, the leading automotive and technology companies in the autonomous mobility space have outlined the importance of their simulation capabilities and have stressed how real-world testing alone is insufficient in developing robust and resilient AVs able to safely and efficiently navigate across the natural driving environment [64].

## IV. THE STATE OF AV SIMULATION

*A. Background*

The complexities of the modern AV pipeline and its integrated subsystems push the boundary on the design and validation processes developed for traditional automotive systems. The integration of DL capabilities into mission-critical safety systems introduces challenges that encompass the development and verification of AV systems, and will require innovative methods that can demonstrate the capabilities of these systems, as well as their safety and reliability in off-nominal conditions. While the specific development and certification standards and requirements are yet to be defined, simulation stands as the key component in the rigorous, scalable, and repeatable design and testing of AV systems.

As presented in [33], there exists a high correlation between cumulative accidents and cumulative autonomous miles. Therefore, a plateau in accidents over an increase in autonomous miles driven, evidenced by more mileage and time between failures, indicates an improvement in the safety of the vehicles. However, this metric alone does not indicate the true robustness of a fully autonomous system, given that the performance is already expected to increase with nominal miles driven over the same roads, same cities, and same scenarios. Instead, simulation techniques and risk extraction methods must be leveraged to expose the AV to critical scenarios never encountered during real-world operation.

Simulation enables the exploration of the probabilistic nature of AV systems and provides a repeatable and rigorous framework for the assessment of their capabilities across a wide range of scenarios. This includes prediction, explanation, and retrodiction of the system's behaviors [65]. These capabilities provide the methodology required to shift the operational risk dynamic such that engineers are able to maximize known and safe behaviors, while minimizing the space of unknown or potentially dangerous behaviors. While rigorous and scalable simulation capabilities are able to shrink the space of dangerous and unknown behaviors, they do not replace real-world testing. Instead, they form a powerful tool that can evaluate the safety of AV systems and their capabilities before they are deployed for real-world testing [34].

With the combined complexity of both the AV system and the environment it operates in, virtual environments cannot emulate every detail, behavior, and interaction from the real world, as then the problem becomes validating the simulator, which is even harder than validating the safety of an autonomous vehicle [39]. Instead, a simulation environment must be able to simplify the system in question, while maintaining the required correlations with reality, so that the behavior of the systems in the simulator can be comprehended as true with respect to the target reality [65]. Therefore, simulation can be conducted across a range of abstractions – from full virtual simulation, where environment and traffic agents are rendered synthetically, to closed-course scenario testing [66], [67], [64], where real-world interactions are simulated using controllable physical props and scenarios. While closed-course scenario testing provides a means to physically simulate the AV across challenging scenarios, many more edge cases can be assessed through the use of virtual "X"-in-the-loop simulation frameworks, where "X" could represent the software, the hardware, the vehicle, or the human [45].

In addition, a high-fidelity simulation platform alone is not the solution – an effective simulation framework is also required to prioritize which scenarios to explore, maximize scenario coverage across AV capabilities and failure modes, and rank failure scenarios in order of importance for further analysis [44]. Furthermore, the driving policies learned in simulation must be able to be effectively transitioned into the target reality and continue to show robust and resilient behavior as they operate and learn in the real world.

The major drivers of the recent developments in AV simulation capabilities span both the academic research as well as the industry pushing towards the deployment of L4 AV technology [17]-[22]. These companies are not only leveraging simulation to teach their AVs to drive safely and legally, but to also teach them how to drive "normally" – mimicking the way that human drivers would navigate uncertain scenarios and interact with other traffic participants. This type of simulation-based learning utilizes the virtual environment's ability of exposing the AV systems to a wide range of edge cases – which ultimately introduces inter and intra-class variations and perturbations across the AV pipeline. Examples of these synthesized variations include enabling AV perception systems to recognize the key traffic signs and their countless structural variations and occlusions [23], and introducing virtual counterfactual scenarios rooted in their real-world observations as part of a replay testing framework [64].

The state of AV simulation can be broken down into two key categories, which when combined, enable the training of robust and resilient systems capable of emerging from the guardrails of simulation and successfully operating in the real-world environment. The first simulation category outlines the ability of simulation to generate the inter and intra-class variations required for learning safe and efficient behaviors across the long tail of driving, even in the face of scenarios with a high degree of uncertainty. The second simulation category captures the work being done to bridge the gap between simulation and reality, enabling the deployment of AV vehicles at scale.

*B. Learning Robust Behaviors in Simulation*

*1. Accelerated Scenario Search Techniques*

Simulation is a critical tool for presenting AV systems with the wide variety of scenarios and the intra-class variations that will be encountered during real-world operation. The synthesized training data representing these challenging scenarios enables the learning of robust and resilient behaviors outside of the constraints of the real world. This goes beyond obeying traffic laws and avoiding collision – it must include non-routine hazards like downed power lines and flooding, as the overall safety of the system hinges on the training and validation data [41]. The probability of encountering these types of scenarios during real-world testing is extremely low and many of the challenging scenarios that an AV could encounter are unknown and constantly changing due to the nature of the driving ecosystem. These edge cases may include challenging interactions with other traffic participants, scenario variances outside of the training scope, and environmental perturbations that push the AV to unexplored behaviors. Therefore, the ability to leverage simulation to expand the types of scenarios an AV system has experienced is a critical component to developing robust AVs able to navigate across the stochastic driving environment.

One of the key challenges to generating interesting scenarios which expose off-nominal AV behaviors is the size and complexity of the input sample space. The real-world environment supports an infinite degree of inter and intra-class variation; therefore, the execution of a wide range of simulation scenarios can be computationally expensive. To overcome this problem, a range of accelerated scenario selection and evaluation techniques have been introduced. These techniques leverage tuned search criteria to generate potentially interesting scenarios and rank the associated risk. In [68] and [35], Zhao et al. present an approach capable of accelerating the scenario generation and AV assessment process, based on the concept of 'accelerated longitudinal evaluation'. This concept is widely used by the automotive industry to predict the robustness of physical components over an extended period of time. In the presented work, they transfer the approach of 'accelerated longitudinal evaluation' to AV assessments, by breaking down difficult real-world scenarios into components that can be repeatedly evaluated through simulation. To capture driving behaviors across real-world scenarios, real-world driving data is collected and distilled down to only the interactions and events of interest. To enhance this type of search problem, they apply importance sampling (IS) to skew the distribution of driving scenarios in a way that increases the number of critical driving events, while maintaining the statistical correctness of the data to ensure it accurately reflects real-world driving situations. They model the driving behavior of the threatening vehicle in the challenging scenarios identified as random variables across a probability distribution, and then run Monte Carlo tests with the accelerated scenarios to expose the AV to a concentrated set of challenging interactions and crashes. This approach effectively replaces the nominal driving conditions with a high-level of exposure to off-nominal safety scenarios, which accelerates the evaluation rate. They apply this methodology to explore the response of an AV to unsafe lane changes and cut-ins from other vehicles – a challenging scenario that must be handled by AVs. Through the proposed acceleration framework, they outperform the limitations of naïve Monte Carlo and are capable of exposing the AV with challenging scenarios in 1000 miles of simulation exposure, equivalent to approximately 2 to 20 million miles of real-world driving.

O'Kelly et al. [34] build on the techniques proposed in [68] and present a larger scale end-to-end AV testing framework capable of simulating rare events using IS techniques. The photo-realistic and physics-based simulator presented is able to feed perceptual inputs, including traffic and environmental conditions captured through video and range data, to the AV system under investigation. The simulator is agnostic to the complexity of the AV driving policy and treats it as a black box model, enabling scalability and expansion across AV systems. In alignment to the work presented in [68], they estimate the baseline human driving model based on standard traffic behavior from USDOT. They then leverage generative adversarial imitation learning models (GAILs) to learn the distribution of human-like behaviors of environmental vehicles. The IS technique is utilized to identify alternative distributions that generate accidents more frequently. In addition to identifying the alternative distributions, they rank them according to their likelihoods under the base distribution, which allows for deeper understanding of the modes in which the AV system can fail and assists in prioritizing the improvement presented to the ego-vehicle policy. As an example of this implementation, an end-to-end highway autopilot network is simulated, which ingests rendered camera images and point-cloud observations from the virtual environment as input and outputs actuation commands. They set up the environmental agents with policies that follow the distribution learned and have a goal of reaching the end of a 2km stretch of road. The IS-based method leveraged learns to frequently sample increasingly rare events and generates 3-10 times as many dangerous scenarios compared to naïve Monte Carlo. They highlight that the acceleration is especially evident when rare event probabilities get smaller.

Norden et al. [44] expand on the work done in [34] by presenting how an accelerated rare-event simulation framework leveraging IS can expose challenging off-nominal conditions to a leading aftermarket ADAS system – Comma AI's OPENPILOT solution [22]. While maintaining a black box relationship with the AV system, the proposed simulation framework is able to assess the ability of the DL-based driving policy to safely interact with human-driving behaviors. To enable the ability to evaluate the AV policy under a data-driven model of the world, they employ a probabilistic risk-based framework that assigns probabilities to environmental states and agent behaviors. The IS method is then utilized to find rare failures, return an unbiased estimate of risk, and rank the scenarios according to the learned probabilistic model. In addition, the scenarios that cause faulty behavior are presented back to the engineering team for prioritization. In addition, they highlight the advantages of this approach to traditional "fuzzing" methods, where slight variations of predetermined scenarios are presented to systems as an attempt to improve the generalization of performance. In comparison, leveraging fuzzing to test AV systems is challenging due to the unbounded nature of the high-dimensional parameter space that the AV system operates in.

In contrast to the importance sampling frameworks, Tuncali et al. [69] propose a falsification approach based on minimizing a robustness metric to accelerate the identification of challenging scenarios. This metric enables the evaluation of simulation outcomes as a cost function, which is able to guide the search of simulation scenarios towards key failures and scenarios of interest. The robustness metric measures how close the trajectory of the system got to an unsafe state, which indicates how close each scenario is to a failure case. The scenario-search function seeks to minimize the robustness value by changing the parameters of the initial states and initial inputs of the system for the next iteration of test case to be run through the simulator, with the objective of identifying the boundary between safe scenarios and collision scenarios. Given that this optimization method seeks to falsify a formal safety requirement, a negative robustness value means the requirement is falsified. By utilizing a distance-based metric instead of a traditional pass or fail flag, they explain how this methodology can handle the inevitable differences between a virtual environment and reality – where some test cases which fail in the simulation may not fail in the real world and vice versa.

As a means of exploring multiple fitness functions, Ben Abdessalem et al. [70] propose a multi-objective search technique that identifies scenarios that stress the system under test. This technique shows how minimization can be conducted across three ADAS fitness functions: (1) a measure of distance between the pedestrian and the collision warning areas in front of a vehicle, (2) the time to collision, and (3) the distance between the pedestrian and the vehicle. They posit that the ability to search across multiple objectives is advantageous in providing the flexibility to explore the critical factors individually, as well as the ability to generate multiple scenarios based on their various interactions. In addition, they leverage a set of surrogate models, which are able to approximate the fitness function values within a set of confidence intervals, as a method to reduce the execution time of computationally expensive simulation processes. The surrogate models are able to estimate the outcome of the fitness function in a significantly shorter period of time than it would take to run the actual simulation. Furthermore, they show how this methodology can be used to evaluate a Pedestrian Detection Vision (PeVi) ADAS system by generating new and challenging scenarios that extract potential AV faults within a limited time budget. In addition, the work presented in [71] shows how multi-object search techniques can be expanded to detect feature interaction failures. These types of failures are some of the key challenges in building robust AV systems due to the fact that these systems are composed of integrated subsystems which tend to impact each other's behavior in unpredictable ways. They cast the problem of detecting the faulty feature interactions as a multi-objective search problem to identify undesirable behavior across many competing objectives. They showcase how this approach can detect real system faults that were unknown to engineers beforehand. Ben Abdessalem et al. [72] expand on this work by combining multi-objective search with decision tree classification models. The generated simulation scenarios are used as input data for building the decision tree, which classifies each generated test scenario as critical or non-critical, based on the simulation output. The decision tree is then employed to generate the set of critical test scenarios and critical regions for the driving policy under investigation. This characterization of the AV input space helps engineers debug their system and identify environmental conditions that are likely to lead to failures.

*2. Scalable Generation and Evaluation of Complex Scenarios*

To enable the ability to perform these types of accelerated risk extraction methodologies in a flexible virtual environment, Best et al. [73] present a high-fidelity simulator able to rapidly piece together complex traffic events and dynamic environmental conditions. These scenarios are capable of supporting various lighting and weather conditions, road hazards and debris, and a mix of traffic participants that place the AV system in dangerous and reactive scenarios. The main objective of this platform is creating scenarios that mimic the erratic, inconsistent, and dangerous behavior of unpredictable actors, and force the AV to react quickly to avoid the hazards presented. They showcase the ability to structure challenging scenarios including passing a bicycle, handling a jaywalking pedestrian, engaging an emergency stop at high speed, navigating high density traffic, and cut-ins. Another high-fidelity simulator proposed by Dosovitsky et al. [74], CARLA, is introduced as a large scale, open urban driving simulator platform that supports the training, prototyping, and validation of AV perception and control systems. This platform can be used to construct challenging scenarios for end-to-end architectures and classic modular pipelines and fills the need for a customizable simulator that can present the complexity of urban driving, including pedestrians, intersections, cross traffic, traffic rules, and other complications to the AV system while supporting high fidelity requirements.

In contrast, Abbas et al. [75] leverage the prebuilt ecosystem that is provided in the video game Grand Theft Auto V (GTA V) to enable high-fidelity assessments of an AV system, without the technical overhead of creating a world simulator which is able to present the AV pipeline with realistic input scenes and challenging traffic and physics. This enables a black-box integration with the AV and allows engineers to focus on selecting the specific weather and traffic conditions, as well as navigation objectives that should be presented to the AV. In addition, they show how this approach is able to compute a measure of how dangerous each simulation was by utilizing a simple measure of minimum distance between AV and other traffic participants. This information is leveraged to initialize the next simulation parameters, including the state of the AV and environmental and traffic conditions, through an optimization routine designed to find dangerous scenarios where the AV gets closest to obstacles. These findings are then returned to engineers to examine the behavior of the vehicle and determine the various causes of the failure, such as missed obstacles, late detection, or an overaggressive controller.

In combining high fidelity scenario generation capabilities and scenario selection, Pei at al. [55] present DeepXplore, a whitebox framework to systematically test DL-based AV systems. To do this, they introduce neuron coverage as a way to assess which parts of a deep neural network (DNN) are exercised by the test scenario presented. This is similar to code coverage in traditional systems, which is a standard measure for how much of code is exercised by an input. By utilizing neuron coverage, they are able to identify specific inputs that achieve high neuron activation and trigger differential behaviors between DL systems through a joint optimization problem that maximizes both components. The inputs DeepXplore generates must remain consistent and realistic to the real world, so the simulation framework applies domain-specific constraints, such as modifying the lighting of an image and adding various layers of occlusions. The simulated scenarios are then used to identify differential behaviors, by leveraging multiple DL systems with similar functionality as cross-referencing oracles to automate the validation process. This allows for the identification of potentially erroneous behaviors without manual labeling and checking. For example, if three different AV policies are tested against a given scenario, and two decide to turn right while one decides to turn left, erroneous behavior is detected between the decision boundaries of the DL systems. They show how this approach can find thousands of incorrect corner case behaviors and explain how the tests generated by this framework can be used to retrain the DL system to improve the accuracy and robustness.

Tian et al. [76] build on the work presented in [55] through DeepTest – a systematic simulation platform that is able to automatically detect erroneous behaviors of DL components within the AV architecture. Using a similar approach as [55], they generate test scenarios by synthesizing real-world changes in the input data representing the addition of rain, fog, and changing lighting conditions. They leverage these translations to systematically explore the different parts of the DNN logic by utilizing the concept of neuron coverage introduced in [55]. They demonstrate how single and stacked image transformation can be used to generate synthetic images that emulate real-world variations and result in the activation of different sets of neurons in the AV DL networks. To further increase coverage, they employ a neuron-coverage-guided greedy search technique which selects combinations of image transformations that increase the neuron coverage metric. In addition, they highlight a systematic way to partition the input space into different equivalence classes based on neuron coverage, where inputs that have the same neuron coverage result in similar behavior for the target DNN and belong to the same equivalence class. Test scenarios are then selected across equivalence classes to provide sufficient coverage. In contrast to the differential behavior approach in [55], they utilize transformation-specific metamorphic relations between multiple executions of the tested DL component to identify erroneous behavior. This technique leverages the fact that regardless of how the driving scenario was synthesized, it is expected that the driving behavior remains consistent between the synthesized image and the original image. For example, the vehicle should behave in the same way for a given scenario under different lighting conditions. This approach results in the detection of thousands of erroneous edge case behaviors without manual specification and intervention.

Zhang et al. [77] expand on the work done in [55] and [76] and present DeepRoad – a simulation platform for automatically assessing DL-based AV systems. They expand on both the fidelity and diversity of the synthetic images to enhance efficacy and reliability, as well increasing the alignment between the training and application domains to further improve accuracy and robustness. Unlike the prior tools, DeepRoad does not use image translation rules to generate the synthetic scenarios based on translations, shears, layers, and rotations. Instead, it is able to produce a much more diverse and realistic set of driving scenes with various weather and light conditions through the use of a generative adversarial network (GAN) techniques. Furthermore, they argue that it is challenging to ascertain if the erroneous driving behavior discovered in the prior tools was absolutely due to the AV system, or if it was due to the inadequacy of the tool to provide a realistic synthetic scenario. Similar to the approach presented in [76], they employ metamorphic testing techniques to validate the consistency of the DNN-based AV system. In addition, they ensure that the synthesized input images align with the application domain, by extracting high-level features and comparing the distances between the training and input images to ensure image similarity.

Expanding on the ability of GAN-based approaches to generate challenging scenarios for learning robust driving policies in simulation, Uricar et al. [78] outline the various applications of GANs in AV development and testing. They highlight its main use as an image-to-image translation technique for style transfer from synthetic to realistic images, or the transfer of style across lighting, weather, and environment domains for AV perception system training and assessment. As a data augmentation tool, they present the prominent work done to create photo-realistic looking images and outline the work being done on context-aware object placement, which enhances existing data sets with realistic looking objects including pedestrians, other vehicles, and road hazards. In addition, they present the various 2D and 3D synthesis frameworks, which enable the generation of new views from a single input image. They also outline the key work done on object detection in the presence of image occlusion. Given that objects are often occluded in real-world driving, the ability to synthesize and present occluded images is essential for training an AV to handle the various detection-based uncertainties that will be encountered during operation.

Wang et al. [79] expand on the importance of being able to generate challenging scenarios which represent the occlusions and deformations present in the real-world. For a safe and capable AV perception system, it must be able to robustly handle illuminations, deformations, occlusions, and other intra-class variations, as any classification errors in this layer have the potential to severely impact all downstream components. They posit that even if a large-scale dataset containing these key scenarios is collected, occlusions and object deformations follow a long-tail and would still be rare scenarios. Therefore, an approach to develop an object detector that is invariant to occlusions and deformations by leveraging GANs is outlined. Instead of generating the synthetic occlusions and deformation images in the pixel space, the adversary executes feature-level changes that mimic real-world occlusions and deformations that are then leveraged to train the object detector. Given that collecting this wide range of rare data is infeasible in reality, GANs are critical components in the generation of realistic data sets that can be leveraged in simulation to develop robust AV performance.

Adding on to the core capabilities of simulation techniques to present a wide range of off-nominal behavior, Rastogi et al. [80] show how a VR simulation platform can be integrated with real-world driving data from a live vehicle to provide flexible scenario-generation capabilities. The VR experience is generated by ingesting data from the real-world vehicle running during the simulation and presenting it to the virtual AV, which updates its position and behavior based on the real vehicle. The AV state is then fed back to a display in the real vehicle to maintain coordination and integration, which allows the simulation platform to reflect live human driving behavior and enables an evaluation of the robustness of the system in real-time. This type of integration is not unique to academia, as the leading AV companies are using integrated simulation environments to generate a wide range of challenging scenarios which can improve the performance of the driving policies learned in simulation. These simulation platforms are tightly integrated with their associated real driving data feeds to enable the exploration and replay of challenging scenarios encountered in real-world operation [64].

### C. Simulation to Real World Transfer

#### 1. Validity of Simulated Outcomes

Given the capabilities of simulation to bypass real-world limitations and expose AVs to the wide range of edge cases required to build robust and resilient autonomous systems, driving policies learned in simulation must be able to be effectively transitioned from simulation into real-world operation. In addition, these driving policies must be capable of continual learning as the AV systems increase in capability and expand across operational domains. This transition is not simple, as AV systems that appear robust in simulation may fail once deployed to the real-world environment, due to differences in appearance and interactions.

Grim et al. highlight how simulations can be labeled as "doomed to succeed", given that they can be used to prove anything [65]. They stress that careful consideration must be taken between the purpose of the simulation and its real-world target, precisely identifying and emphasizing the aspects that are intended to correspond to reality and those that are not intended or required to do so. In other words, simulations don't have to be nearly identical models with all elements represented, but rather abstractions that maintain the correspondence of the essential or critical elements. To further expand on this, the notion of 'intentional non-correspondence' is developed to emphasize the distinction between the aspects of the simulation that were intended to correspond to reality and those that were not. For a successful simulation, it must be capable of objectively sustaining the correspondence between key aspects of the simulation's structure and known aspects of reality, providing the ability to analyze other aspects of the simulation structure as corresponding to unknown aspects of reality. When the structure of intended correspondence and intentional non-correspondence breaks down, the simulation results will not produce an output that corresponds to expected reality, rather mere artifacts of the simulation process. In addition, they highlight that correspondence at the point of new information is the crucial element for a simulation to successfully achieve its purpose; therefore, if failure occurs with new but relevant information, then this is a clear and dramatic simulation failure. They also note the importance of ensuring that the simulation does not commit a sin of commission, in which the simulation involves features intended to correspond to reality but do not, or a sin of omission, in which crucial aspects of the target reality are excluded from the simulation.

Abbas et al. [75] expand on this by posing a fundamental question about the ability to transfer AV policies learned in simulation to reality – do accidents that are discovered in simulation reveal something about how the AV would perform in the real-world? Given that it is impossible to obtain a direct empirical answer to this question without running the same scenarios in the real world, two proxy questions are proposed that can be answered instead and help establish confidence in simulated results: (1) "what is the relation between the perception algorithms' performance on synthetic driving scenes rendered by the graphics engine and their performance on natural driving scenes?" and (2) "what is the relation between the effect of the AV controller's actions in the simulator and their effect in the real world?". These two questions cover both the perception pipeline validity, as well as the validity of the controller and dynamical model of the vehicle. Abbas et al. state that if they have confidence in the answers to these two questions, then they can have confidence in the usefulness of the AV simulation and testing framework. Furthermore, they present an approach which focuses on the first question, given that there are standard dynamical models which are accepted by the automotive industry and thoroughly validated by automotive engineers to support the second question. To compare the performance between perception algorithms on synthetic and natural scenes, they compare the object detection performance and the visual complexity between prominent synthetic and natural driving data sets. Their findings are that the synthetic scenes generated in simulation produce a lower recall and thus present conservative results that would not mislead the engineers to overestimate the capabilities of an AV system. In terms of visual complexity, they compare key metrics to estimate the density and distribution of edges, textures, colorfulness and contrast variations across the two sets of images – finding that the simulated world results in intermediate complexity which sufficiently overlaps with the real scenes. These findings support the validity and usefulness of simulation-based training with respect to the natural driving environment.

#### 2. Perturbations and Randomization in Simulation

To enhance the ability of the simulation-learned driving policies to effectively transition to real-world operation, multiple methods have been proposed. Tobin et al. [81] present a domain randomization approach to bridge the reality gap that separates simulation and the real-world environment. They show how models trained on simulated images can effectively transfer to the real-world environment by randomizing the simulator inputs experienced during training. Given a high degree of randomness in the simulator, the natural environment would appear to the model as another variation of the random inputs it was trained on and it would be able to generalize to the real-world with no additional training. By leveraging domain randomization, they are not constrained to using high-quality renderings in the simulation, which match the exact textures, lighting, and configurations in the real world. Instead, by increasing the variability of the simulator they are able to optimize for speed and lower the cost of achieving transferability from sim to real. While their work was focused on the robotics space, it has been employed across AV simulation work. [49] and [50] leverage domain randomization to introduce perturbations and disturbances in vehicle trajectories and positions during simulation training, which improved the performance of their end-to-end driving policies once transferred from simulation training to real-world operation. As part of the training of their imitation learning agent, Bansal et al. [51] also leverage synthesized perturbations to the expert driving trajectory, which expose the learning agent to potential collisions and off-road behaviors. They highlight the minimal weight given to such perturbations, so that they are sufficient in frequency to experience the benefits, but the agent does not learn a propensity for perturbed driving.

In addition to injecting randomness to the simulated domain, Zhang et al. [82] highlight the importance of leveraging real-world image sequences integrated with road GIS data to render the virtual realities within the simulation environment. They propose a simulator which enables the AV driving policy to be trained on actual images turned into virtual scenes, ultimately improving the transferability and stability of the driving policy to the real world. Their proposed RoadView simulator provides a more photorealistic scene compared to just employing simulation engines that render virtual realities. The work presented by Amini et al. [83] expands on the types of photorealistic scenarios generated within simulation, by synthesizing off-orientation virtual scenes based on real-driving data. These scenes force the vehicle to conduct the off-orientation recoveries that could be demanded during real-world operation. To do this, they present a simulation and training engine, VISTA, that is able to train reinforcement learning-based AV agents, entirely in simulation, without any prior knowledge of human driving or post-training fine tuning. The VISTA engine allows the virtual AV agent to not only learn from sensory data based on stable human driving, but also from an expanded set of simulation-generated data representing near-crash scenarios on the road. This is done while maintaining the fidelity and semantics of the real-world environment. These agents are able to be deployed directly into the real world and navigate roads and environments not encountered before. They argue that this expansion of training AVs beyond the parameters of traditional imitation learning enables a scalable method to learn policies that can transfer to and navigate in previously unseen environments and complex, near-crash situations. They show that this approach exhibits greater robustness in the real-world and enhanced recovery frequencies when compared to standard IL driving policies and domain randomization techniques. In addition, Zhang et al. [77] leverage a GAN-based approach to synthesize diverse driving scene representations of challenging weather conditions like snow, rain, and fog. These scenes characteristically align with the real-world driving environment and improve the transferability of driving policies in the presence of challenging real-world weather conditions.

Uricar et al. [78] survey the various techniques and methodologies in which GANs are leveraged to generate synthesized images that better align with the real world and have the potential to introduce unique perturbations into the simulated environment. In the survey, they highlight the challenges with using simulated environments to train agents, which often fail to generalize on real environments. Furthermore, they present the work being done in domain adaptation to enhance the ability of DL-based autonomous driving policies, that are trained in the simulation domain, to generalize to the target real-world domain. In addition, they outline the abilities of GANs to improve object detection capabilities, especially for scenarios of object occlusion, which occur frequently in the real world but must be synthesized in the simulated world. They also present the inpainting capabilities of GANs, which enable the handling of noisy or incomplete readings that sensors may suffer in real world operation. As a specific application of GAN-based image transformations, [84] introduces sensor-level perturbations and occlusions that mimic the environmental conditions in which some AV applications may need to operate in. This work presents how soiling (mud, rain drops, freeze, dust) and adverse weather conditions (heavy rain, blizzards) can cause deterioration of sensor performance and image recognition capabilities. To generate these scenarios in simulation, they employ a GAN, where the generator learns to introduce realistic soiling patterns to an image as well as the corresponding annotation masks. This allows the simulation of sensor-level perturbations and showcases the importance of being able to generate this type of data to ensure the sensors and perception system are able to handle these expected scenarios in real world operation.

Furthermore, Chalaki et al. [66] show how perturbations in simulation can be expanded beyond ego-vehicle driving policy training to an ecosystem of connected and automated vehicles (CAVs) – where adversarial perturbation impact both input and output of AVs attempting to coordinate. To demonstrate, two sets of RL experiments are conducted, where one policy is trained on Gaussian noise injected into state and action space, and the other policy leverages a second adversarial agent with a reward function incentivized by the learning agent's failure. In the adversarial case, the second agent attempts to perturb elements of the learner's action and state space, including observed positions, velocities, and distances, while the learner attempts to coordinate with other agents to cross a roundabout. In both the Gaussian single-agent and adversarial multi-agent experiments, the driving policies learn how to execute ramp-metering actions, where one group of vehicles slow down to allow the other group to pass. The adversarial policy outperforms both the human-driving baseline and adversary-free training and demonstrates improved performance compared to Gaussian noise injection after transfer to the real-world testbed, evidenced by reduced travel time and increased average speed. This outlines a zero-shot sim to real transfer that utilized adversarial noise to improve the performance of integrated CAV maneuvers under stochastic, real-world disturbance.

*3. Translation and Integration across Simulation and Reality*

Pan et al. [85] present a realistic translation network that converts non-realistic virtual images into realistic ones that mimic the scene structure of the target real world, using similarities between the segmented scene structures. Even though the virtual images have different visual details than real driving scenes, they share a similar scene parsing structure given that both sets of scenes share similarities such as roads, foliage, buildings, and other vehicles. Given this shared space, the team's method consists of translating from virtual to realistic images employing the scene parsing representation as the interim step. To do this, they leverage two modules: one that converts virtual scenes to the parsed representation, and the other module that converts from the parsed representation into the realistic images. This scene parsing transformation frees the translation process from being reliant on existing mappings between virtual images and real images. Therefore, by using the transformed realistic images, the driving policy is able to learn on realistic images and enhances the transferability from simulation to operation in the real-world environment. They show that this method has the potential to bridge the gap between simulation and reality. In comparison, Bewley et al. [86] present a modified approach that develops a translation mapping between virtual and real images while jointly learning a vision-based driving policy from a common latent space, representing the common structures between simulation and reality. This is used instead of more explicit representation like semantic segmentation [85].

As another technique to bridge the differences between reality and simulation, Muller et al. [38] leverage modularity and abstraction to encapsulate the driving policy in simulation and enable a more controlled transition to reality. This encapsulation between upstream and downstream components prevents the driving policy from directly interacting with raw perceptual inputs and generating low-level vehicle controls, as these direct interactions could introduce the areas of significant variance between simulation and reality. Architecturally, it provides the modularity required so that the perception system generates a semantic map, which is then used by the driving policy to generate a planned trajectory defined by a series of waypoints. The waypoints are provided to the low-level motion controllers which guide the vehicle along the specified path. While the driving policy is purely trained in simulation, it is trained on the output of the actual perception system, and not perfectly labeled and segmented ground-truth data. This allows the driving policy to adapt to the perception system's imperfections. The research group demonstrate the capabilities of this hybrid end-to-end system by showing how driving policies transferred across different simulated weather and environment conditions outperform traditional end-to-end methods. They also show the capability to transfer the driving policy from simulation into the real-world leveraging a 1/5-scale robotic truck, which is able to effectively navigate across a variety of roads and environmental conditions. This type of translation and transfer approach is more generalizable in comparison to [87], which develops a highly engineered 3D simulator to mimic the target domain. In comparison to [85] and [86], the creation of a 3D environment for each setting leads to significant engineering challenges with scalability.

In addition, Chao et al. [88] highlight the importance of integrating realistic mutual interactions into the simulated environment. While this is a critical component in ensuring the validity of driving policies trained on human interactions, it still remains a key gap in AV simulation. The objectives and interactions of the pedestrian-traffic that must be modeled can take multiple forms in an urban environment, therefore they leverage a "force-based" concept to build a heterogenous traffic simulator with the goal of accurately representing these human behaviors and interactions. The "force-based" concept enables the participants in mixed traffic to act as if they would be subjected to repulsion and interaction forces based on their objectives, their neighboring participants, and any static environment features. In comparison to leading AV simulators like Apollo [89] and Carla [74], which leverage pre-defined behaviors of non-vehicle road users, the approach presented by Chao et al. attempts to solve the problem of considering the mutual influences and real interactions between vehicles and other human road-users. To provide a comparison, they present a scenario where a pedestrian is crossing the road with an AV approaching. In the Apollo simulator, the pedestrian continues crossing the road at a predefined velocity and does not consider the behavior of the incoming vehicle. In the proposed force-based approach, the pedestrian and the vehicle interact with each other and change their trajectories based on each other's instantaneous states. This type of mutual interaction is critical in transferring simulation learning that enable an AV to navigate complex intersections that require mutual interaction. They construct this interaction simulation framework in such a way that an AV driving policy can be plugged into the simulator environment and utilize its own sensing and planning systems to interact with pedestrians, bicyclists, and other traffic participants in the system in a more realistic fashion.

Li et al. [90] present a solution to further expand the interactive behavior presented in [88] by introducing a decision-making process that reflects human behaviors and interactions. To simulate the actions of human drivers, a decision-making process is integrated into the simulation, instead of a traditional approach of utilizing a set of prescribed actions that are based on a function of time or the state of the system. To do this, they employ a game theory-based approach to develop a traffic model that enables all the vehicles in the presented scenario to simultaneously act as decision makers, emulating the mutual interactions experienced in the real-world. The integration of a hierarchical decision-making approach is based on the fact that intelligent agents, like drivers, have different levels of reasoning, represented by a 'level-k' driving model. These agents are assigned different levels, where a level-0 agent does not consider the probable actions of other agents during the interaction, limiting their behaviors to be reflexive and reactive. For example, a level-0 driver will react by braking hard if they observe an obstacle in the road, without considering how the car following them would behave given their sudden deceleration. In contrast, presented with this same obstacle, a level-1 driver would consider that the car following them is being driven by a level-0 driver who would react by braking hard. When this is not enough to avoid a collision from behind, the level-1 driver would choose to make a lane change to avoid the impending collision. Expanding on this hierarchy, a level-2 driver would assume that the other drivers are level-1 and would act based on the expected behaviors of the other drivers. To obtain these level-k driver policies, a training process is followed, where the trained driver is defined as the "learner" and the other vehicles constitute the "environment". The training process starts with a level-0 driving policy, where drivers do not take into account interactions with the other drivers and do not consider their potential behaviors. The level-1 policy is then trained through a reinforcement learning algorithm, where the environment is assigned to level-0 behavior. Similarly, the level-2 policy is learned where the environment is assigned level-1 behavior. While this hierarchical assignment can be expanded to higher levels, the authors decide to train the policy up to and including level-2 given that this level of interaction and behavior planning best represents human driving interactions. Furthermore, they show how AV control algorithms can be quantitatively evaluated through this framework to optimize the outcomes with respect to the behaviors in the real-world environment and leverage these interactive scenarios to calibrate the parameter values in an AV control algorithm to better reflect reality and improve the performance once transferred from simulation.

In addition to ego-vehicle simulation, Li et al. [91] outline how a tight integration between field-vehicle operations and simulation improves simulation capabilities and overall system performance. They present a parallel testing system that enables the system to continue collecting real-world data through the field-vehicle, while updating the simulation with the new data. In the opposite direction, the simulation outcomes guide the field tests. Schluse et al. [92] expand on this parallel integration by introducing the hybrid application scenario in which experimentable digital twins (EDT) are used in combination with physical systems in real-world operation and not just limited as engineering and testing tools. This integration enables the execution of complex planning, prediction and control processes in intelligent systems, creating the real-time synchronization between a physical system and its corresponding EDT. The proposed integration provides a seamless transition between virtual and physical systems. They highlight that the ability for simulation outputs to integrate into physical systems, operating in the real-world, is a key enabler of the development of intelligent systems. Specifically, this facilitates the development of complex control algorithms and their transfer to the real system, while enabling the realization of "safe systems" that are able to supervise themselves and evaluate decision alternatives using an EDT in simulation.

*4. Continual Learning*

As an extension of the key task of transferring simulation-based learning into the real-world environment, continual learning is a critical ability for AVs. They must be able to continue to learn new tasks without having the new learning interfere with the old learning – this is called "catastrophic forgetting" [93]. If a DL system must be retrained every time a new task or update is introduced, it would limit the ability to scale and deploy autonomous vehicles. In addition, there is the opportunity to leverage the transferability between simulation and the real-world, by utilizing the simulation training as an accelerator to enable continual learning of robust real-world behaviors. While there is the approach of replaying old training data mixed with the new data, it is still not an optimized method for incremental and continual learning of driving policies. Human drivers have the ability to incrementally expand their driving ability across new tasks and scenarios, so AVs should strive for the same ability.

Some recent work in the field of continual learning employs elements of regularization [93], [94], and network expansion [95]. Li and Hoiem [94] present a method that updates the network parameters so that the new task is able to be performed given new data, but also considers the output of the old network on the new data, to minimize changes to the existing learned parameters. This is done by minimizing the loss between the output of the old task and old network on new data, and the output from the updated network. In addition, they leverage a parameter to quantify the bias towards plasticity of learning new tasks and the stability of maintaining performance on old tasks.

Kirkpatrick et al. [93] propose a different methodology called "Elastic Weight Consolidation" (EWC) to solve the problem of catastrophic forgetting. This proposed method enables continual learning by identifying the network weights that were important to the initial tasks and penalizes updates to these weights while training a new task. The goal of EWC is stay in a low error region for the original tasks, while enabling the successful expansion to a new task. This approach also leverages a parameter to balance between plasticity of the network and stability. In addition, [95] introduces the method of progressive neural networks, which utilizes a dynamic network architecture so that new neurons, connections, and outputs are added for each new task, without modifying the weights trained from the old tasks. This approach utilizes the training done so far and bridges the intermediate network layers from the initial network to the expanded tier. With this expanded architecture, scalability is a challenge as more tiers of neurons are added with more tasks learned. Furthermore, [96] introduces a methodology called generative replay, where a generative model is employed to produce synthetic data that approximates the distribution of the original training data. This synthetic dataset is then combined with the new task data so that the network does not forget what is has already been trained on.

The balance of optimization for new tasks will fall between the flexibility to learn new tasks and the stability to maintain performance on old tasks. As this is a space that is still currently being explored, it is an interesting area in which these new learning tasks can be effectively trained in simulation and reliably transferred into the real-world to facilitate the expansion and growth of AV capabilities.

## V. DISCUSSION

*A. Open Problems and Challenges*

As outlined in Sections III and IV, significant effort across disciplines and organizations has been applied to facilitate the realization of robust and resilient autonomous vehicles at scale. The simulation-based development and certification efforts surveyed in this paper have played a key role in the deployment of perception, planning, and control systems capable of successfully guiding AVs through dense and challenging urban environments at a level of performance which builds the trust required to deploy them in limited-domain trials.

While these autonomous driving technologies have come a long way from their initial proofs-of-concept, challenging obstacles still remain in the road towards rigorous training of robust driving policies and the ability to ensure resilient AV operations at scale. Given the complexity of these systems, it is unlikely that a single method will emerge as the oracle for safe and effective AVs [34]. Instead, an expansion and integration across simulation capabilities is required to solve the key challenges that have emerged in the development and certification of AV driving policies.

*1. Defining Vehicle Intelligence*

One of the critical gaps in the evaluation of AVs is that we currently lack a clear and measurable metric for the "intelligence" of AVs [54]. Given that this intelligence represents the capabilities and limitations of the integrated DL-based systems that power AVs, this lack of definition challenges the ability to empirically prove that an AV is capable of driving in live traffic, reacting to human behaviors, and handling scenarios which it has not yet been exposed to.

While there have been efforts to solve this problem through a proxy measure of intelligence using formal verification methods [39], the lack of a definition stands as a critical gap in understanding the capabilities and limitations of AV systems. In addition, the inability to directly define a vehicle's intelligence forces the use of statics-based safety determinations that may not always be supported by sufficient exposure to challenging real-world scenarios [61]. Both scenario-based and functionality-based testing aim to present a wide range of verification scenarios to assess a vehicle's intelligence, but the potential of hidden risks outside of the validation data remains. These proxy measurements of AV intelligence all come with their own risks and caveats; therefore, a measurable definition of AV intelligence is a major challenge to the development and certification of AVs.

*2. Proving Simulator Fidelity and Outcome Validity*

While simulators have shown the capability to expose AV systems to a wide range of off-nominal scenarios which stress the system and effectively extract hidden risks and faulty behaviors, these outcomes are useless unless the simulator can be baselined as a trusted and representative approximation of the target reality. The criticality of leveraging simulation methodologies has been emphasized across federal and industry guidelines and simulations platforms have become a core component of AV policy training – yet simulation can still be charged as "doomed to succeed" and capable of proving anything [65]. Therefore, to enable the effective use of simulation, there are key challenges in fidelity and validity that must be solved for. These challenges include the decisions and considerations taken during simulator development to ensure that the simulator meets the level of correspondence and intentional non-correspondence required, while avoiding the sins of omission and commission [65].

In other words, simulations don't have to be nearly identical models with all elements represented, as their purpose is to provide an abstraction that maintains the correspondence of the essential or critical elements. Therefore, the degree of fidelity of a given simulation is one of the key abstraction challenges. Some platforms leverage graphics engines to render a fully virtual scene [73], [74], [89], and others incorporate the use of real-world and DL-generated images [33], [77], [82] to attempt to align with the required correspondence levels. In addition, [39] underscores the challenge of validating a simulator and argues that ensuring that the simulator faithfully represents the richness and complexity of the real-world environment is harder than validating the driving policy itself.

Furthermore, in comparison to other simulation applications [97], the AV industry lacks a standard simulation framework that has been benchmarked as a valid source of generating simulated outcomes for real-world comparisons. Therefore, ensuring the fidelity and the overall validity of the simulation platform remains as a key challenge to solve in enabling the full utilization of simulation capabilities.

*3. Handling Black-box Interactions*

Most AV systems developed in industry are black-box in nature to preserve the integrity and confidentiality of the intellectual property [44]. While understandable, this presents a significant challenge to anyone trying to evaluate or certify the systems from the outside looking in. Therefore, the integration of black-box systems with simulators is limited to identifying inputs that trigger incorrect outputs. Furthermore, this shrinks the scope of behavioral information able to be extracted from the system and restricts the evaluation to an approximation of probable system and subsystem failures. Given the key role that assessment and certification capabilities will play as AVs scale, the approach to effectively validate a black-box AV driving policy is a key challenge that will need to be solved to ensure transparency and understanding of AV systems.

*4. Ensuring Safe and Efficient Behaviors*

Simulation frameworks have shown their potential to accelerate the selection of test cases and optimize the generation of synthetic data sets representing challenging scenarios, outside of what is possible in the limitations of real-world testing. The identification of the breadth and depth of test scenarios required to provide coverage across an AV system's capabilities and the various environments it will be required to operate in, is a challenging task in optimization and approximation, given that not every possible scenario can be identified and tested [44]. Therefore, simulation frameworks are utilized to select high-priority test cases that extract hidden risks and expose surprising behaviors in an efficient manner. Even so, passing an optimized number of critical test cases does not guarantee generalizability of AV system performance in the real world. Furthermore, given the long-tailed nature of the driving environment, there are many unknown scenarios which could lead to faulty and potentially dangerous behavior. In addition, methods that leverage traditional fuzzing techniques, where small variations are added to predetermined scenarios to improve general performance, is inadequate in the AV space [98], as this method works best where the input space is bounded.

Various approaches have been utilized to optimize and accelerate the scenario selection process [34], [35], [69], [70], [82], but the challenge of finding all high-likelihood failure scenarios remains. This can be thought of as identifying the area which falls below an acceptable threshold across the performance landscape. Capturing and understanding this area is required to fully understand the risk of AVs and accurately estimate success and failure metrics. Expanding on this challenging task, Shalev-Shwartz et al. [39] stress the difficulty of utilizing a purely offline simulator for identifying the areas of potential risk and suggest an integration of a formal model of safety and performance. This shifts the challenge of identifying the testing required to guarantee that the AV will never be involved in an accident, to an objective of guaranteeing that the AV will be careful enough so that it is never part of the cause of an accident.

Furthermore, the faults presented in [25] and [26] expose an additional layer of challenges that must be handled in the simulation and validation of DL-based perception systems. Szegedy et al. [26] show how a perception network can be tricked into misclassifying an image by applying an unnoticeable perturbation to the image. They show that the nature of those perturbations is not due to random artifacts of learning, but rather, the same perturbation can cause different networks trained on different subsets of data to misclassify the same input. This raises a challenging question – if a network is able to generalize well, then how can it be confused by these indistinguishable adversarial negatives? Additionally, Nguyen et al. [25] expand on this to show that they could also generate images that are completely unrecognizable to the human eye, in the form of white noise, yet are able to fool state-of-the-art DNNs to believe they are recognizable objects with 99.99% confidence. Aside from the structural challenges of ensuring safe and efficienct behaviors across the scale of the sample input space, the risks outlined in [25] and [26] expand the complex nature of this challenge by presenting the underlying vulnerabilities of DL-based systems which must be handled in the deployment of AVs.

*5. Modeling Human Interactions*

The ability to emulate human behaviors is one of the critical reality gaps in AV simulation [88]. Modeling human interaction across various traffic-participants is an extremely challenging task. This interaction gap presents a key challenge in ensuring AVs are able to navigate scenarios of higher complexity and higher interdependency with the instantaneous states of the other human agents in the scenario. It is essential that AV systems get exposure to this type of interactive behavior so that they are able to handle complex scenarios, including predicting human behaviors, nudging to gain position, and negotiating to ensure efficient and safe navigation.

Given the unpredictable nature of human behavior, these interactions are typically approximated in simulation through predefined paths or scripted and limited interactions. While [88] presented a 'force-based' approach for emulating human behavior and [90] presented a hierarchical decision-making model that better reflects reality, this remains a key challenge and one of the simulation focus areas for leading organizations in the AV space.

*6. Bridging the Reality Gap*

To enable the use of simulation-based learning, the gap between simulation and reality must be bridged. In [83], the importance of transferring policies learned in simulation into the real-world operation is posed as one of the critical research challenges. Without this, simulation-based driving policies that appear to successfully operate in the virtual environment would quickly fail once exposed to reality, due to the differences between the virtual and natural environment.

In [85] and [86], an approach was proposed to shrink the reality gap by employing translation mappings able to take advantage of the commonalities between virtual and real images. In addition, [81] and [83] outline different methodologies consisting of domain randomization and perturbations to improve the generalizability of the driving policy and its performance once transferred to the real world. Yet, research on this subject is still in its early stages and key challenges remain in facilitating a zero-shot transfer of a complex AV driving policy between simulation and reality. The challenges of this type of translation grow as both the scene complexity and the variety of the associated agents and objects increase. Furthermore, shrinking of the reality gap enhances the potential to fully leverage simulation as a scalable and efficient tool in building the robust driving policies capable of handling new scenarios.

*7. Enabling Adaptivity, Efficiency, and Scalability*

Adaptivity, efficiency, and scalability are key characteristics of AV systems that govern their adoption into our society. Given that these systems are expected to undergo regular updates as well as bug fixes, one of the key adaptivity and efficiency challenges is ensuring that modifications to the code base do not cause other safety-critical components to deteriorate, thus preventing catastrophic forgetting [93]. These modifications can come in the form of enhancements where the AV system is trained on a new task or capability, or in the form of critical fixes that must quickly be implemented given the potential severity of the identified fault. Therefore, it is essential to have an efficient simulation framework that can benchmark current performance against previous performance, identify behavior changes due to updates, and adapt at the pace of development [44]. In addition, it is essential to understand how a single line of code change will be regression tested. A key example of this challenge is presented in [39], where they question if the validation process is efficient and adaptive enough to identify if a single code change has resulted in a new failure that was not present in the validation miles traveled before, or if a full revalidation would be required to identify the impact.

In terms of scalability, Shalev-Shwartz et al. [39] stress that the premise of autonomous driving is more than just "building a better world", rather, that the mobility provided by an autonomous vehicle can be sustained at a lower cost than with a driver. This highlights the importance of scalability, where engineering solutions that lead to unleashed costs will not be able to scale to millions of vehicles. Therefore, along with the challenge of minimizing the cost of computing and sensing, the cost of validation is also critical to the value proposition of AVs, as it enables their scalability across new operational domains.

*B. Potential Research Opportunities*

While open problems and prominent challenges in the AV space range in scale from ego-vehicle capabilities to end-to-end integration and connectivity, three key areas of AV research opportunities emerge, which seek to bridge the challenges and deliver a roadmap to future solutions. These areas of research are: (1) Defining and Measuring AV Intelligence, (2) Enhancing AV Simulation Frameworks and Methodologies, and (3) Advancing AV Simulation Transferability and Integration.

*1. Defining and Measuring AV Intelligence and Safety*

The lack of a clear definition for the intelligence of AVs [54] presents a research opportunity in identifying novel ways of defining and measuring AV intelligence through simulation-based assessments. Aside from the standard method of leveraging a statistical-safety argument as a proxy for intelligence, there are promising areas of research that can be explored including proving generalizability across different scenarios, formal validation frameworks that expand on the concept of "driving carefully" [39], and probabilistic representations [99] of perceived risks.

In addition, the intelligence of a vehicle can potentially be integrated as measure of its ability to understand the dynamics, risks, and diverse objectives and intentions present in a scene, rather than just its ability to map the scene. Furthermore, advancements in the capabilities required to more effectively measure AV intelligence will present additional opportunities in defining unique verification and validation frameworks that rely more on generalizability of an AV system's performance, rather than on defining specific sets of prioritized assessment scenarios.

*2. Enhancing AV Simulation Frameworks and Methodologies*

Given that simulation of AVs will not be able to capture every nuance of real-world driving [80], it is critical to continue to expand simulation capabilities as a means of enabling more generalizable driving policies that exhibit the required invariance in the presence of a constantly changing environment. Therefore, key opportunities exist in increasing the inter and intra-class variability and complexity of the synthesized training images through GAN-based approaches. From a data perspective, this work could facilitate the expansion of existing AV data sets [100]-[105] to include additional complexities that could be utilized to enhance the robustness of simulation-based driving policies.

Furthermore, the expansion of the human behavior simulation work presented in [54], [88], would enable the integration of more realistic human-interaction models, capable of emulating real-world human behaviors in simulation and facilitating the exposure of these types of complex interactions to AVs during the training and validation process.

In addition, while there has been existing work in leveraging agent-based modeling (ABM) to understand AV deployments at the ecosystem level, it has been mostly focused on fleet analysis and ridesharing [106]. This type of modeling and simulation work can be expanded to further explore potential emergent behaviors that could appear across a range of AV interaction and coordination scenarios.

*3. Advancing AV Simulation Transferability and Integration*

To fully utilize the capabilities of AV simulation in the design and testing of autonomous driving policies, the foundational work in scene translation, domain randomization, and tuned perturbations must be expanded to support higher degrees of complexity and an increase in fidelity to the natural environment. There are unique research opportunities to enhance the synthetic representations of occlusions, deformations, and reflections though the use of GAN-based techniques [78]. This type of research will help shrink the gap between synthesized driving frames and the real-world driving environment, enabling the transferability of more robust driving policies.

In alignment with the recent advances in continual learning, there are key research problems focused on enhancing the flexibility of DL-based systems in learning new tasks while balancing their stability to maintain performance on old tasks. Furthermore, this presents an opportunity to create a simulation framework able to facilitate this type of flexible learning by benchmarking current performance against previous performance, identifying behavior changes due to driving policy updates, and adapting at the pace of development to support the transparency required for AV system updates, modifications, and fixes.

Additionally, there are multiple research opportunities aligned to integrated and connected AV systems. A promising area of research that can be explored is the utilization of parallel simulations integrated with live AVs. This integration would facilitate the development of complex mission-safing procedures and the possibility of creating self-supervising systems able to evaluate their behaviors across multiple potential outcomes and assess alternative decision in real-time [92]. As presented in [33], most of the accidents in the AV driving studies conducted so far involve AVs being rear-ended by a conventional vehicle. Additionally, most disengagements were due to the human driver opting for a different form of evasive action. Therefore, there is an opportunity to further mature the live decision-making process of an AV system through an integrated and parallel simulation, with a focus on optimizing human interactions and evasive maneuvers in the face of uncertainty.

## VI. Conclusion

The integration of AV technology into our society is reshaping the future of mobility. These systems are not just an expansion of existing automotive architectures and technologies, but rather, a rewrite of the driving task in which the responsibility to perform in a safe and efficient manner is transitioned from human to machine. These advancements aim to answer the shifting question of automotive safety, as the mechanical and technological concerns posed in the '60s have now transitioned to the limitations and imperfections of human drivers. While AV systems have shown the capability to mimic human driving and surpass human limitations, new risks emerge that must be understood and handled to enable the deployment of robust and resilient AVs at scale.

Given the limitations of real-world testing, AV simulation stands as the critical component in exposing AV systems to the long-tail of real-world scenarios, while enabling the development of perception, planning, and control systems capable of handling the complex interactions and uncertainties that will be encountered in real-world operation. This paper provides a background of the AV industry and outlines the associated development and testing guidelines. Furthermore, it presents a survey of the various simulation frameworks and methodologies that have been developed to explore the edge of autonomous driving capabilities, which facilitate the development of robust driving policies and extract hidden risks to build the resiliency required for real-world operation.

Although significant progress in the AV space has been made in the past two decades, there are still key challenges that must be solved to enable the deployment of AVs at scale. These challenges can be synthesized as: (1) Defining AV intelligence, (2) Proving simulator fidelity and outcome validity, (3) Handling black-box interactions, (4) Ensuring safe and efficient behaviors, (5) Modeling human interactions, (6) Bridging the reality gap, and (7) Enabling adaptivity, efficiency, and scalability.

Furthermore, by bridging the commonalities between the open problems and challenges in the AV space, three key areas of AV research opportunities emerge: (1) Defining and Measuring AV Intelligence, (2) Enhancing AV Simulation Frameworks and Methodologies, and (3) Advancing AV Simulation Transferability and Integration.

Given the scale of the AV ecosystem and the speed at which advancements are emerging, these research opportunities provide a high-level roadmap to guide the development of solutions that enable AV technologies to exit the guardrails of simulation and deliver robust and resilient operations at scale.